\documentclass{article}
\usepackage{spconf,amsmath,epsfig}

\usepackage{subcaption}
\usepackage{comment}
\usepackage{setspace}
\usepackage{enumitem}

\usepackage[hang,flushmargin]{footmisc}
\usepackage[ruled,vlined]{algorithm2e}

\usepackage{hyperref}
\hypersetup{
	colorlinks=true,
	linkcolor=blue,
	filecolor=magenta,      
	urlcolor=cyan,
}

\graphicspath{{figs/}} 

\newcommand{\f}{\mathbf{f}}
\newcommand{\fh}{\hat{\mathbf{f}}}
\newcommand{\x}{\mathbf{x}}

\newcommand{\gt}{\mathrm{gt}}

\usepackage{color}
\usepackage[dvipsnames]{xcolor}

\title{Noisy Supervision for Correcting Misaligned Cadaster Maps\\ Without Perfect Ground Truth Data}
\name{Nicolas Girard$^1$, Guillaume Charpiat$^2$ and Yuliya Tarabalka$^{1,3}$ \thanks{
The authors would like to thank ANR for funding
the study.} }
\address{
	$^1$TITANE team, Inria, Universit\'e C\^{o}te d'Azur, France \\
	$^2$TAU team, Inria Saclay, LRI, Universit\'e Paris-Sud, France \\
	$^3$LuxCarta Technology \\
	email: firstname.lastname@inria.fr
}
\let\OLDthebibliography\thebibliography
\renewcommand\thebibliography[1]{
  \OLDthebibliography{#1}
  \setlength{\parskip}{0pt}
  \setlength{\itemsep}{5pt}
}

\newcommand{\smallestsection}[1]{\medskip\noindent\textbf{#1}}

\begin{document}
%
\maketitle
\begin{abstract}
	
In machine learning the best performance on a certain task is achieved by fully supervised methods when perfect ground truth labels are available. However, labels are often noisy, especially in remote sensing where manually curated public datasets are rare. We study the multi-modal cadaster map alignment problem for which available annotations are misaligned polygons, resulting in noisy supervision. We subsequently set up a multiple-rounds training scheme which corrects the ground truth annotations at each round to better train the model at the next round. We show that it is possible to reduce the noise of the dataset by iteratively training a better alignment model to correct the annotation alignment.

\end{abstract}

\section{Introduction}

One of the main tasks in remote sensing is semantic segmentation. The supervised approach needs good ground truth annotations most often in the form of class-labeled polygons outlining objects of the image. However, these good annotations are hard to come by because even if they exist (for example OpenStreetMap (OSM) annotations~\cite{osm}), they can be misaligned due to human error, imprecise digital terrain model or simply a lack of precision of the original data (scanned cadaster maps from local authorities). Each object can be misaligned in a different way compared to surrounding objects and the misalignment can include complex deformations such as slight stretching and rotation.

The aim of this paper is to correct the alignment of noisy OSM annotations when only these annotations are available. Several related works tackle the noisy label problem. Some use special losses to explicitly model the label noise~\cite{Mnih:2012:LLA:3042573.3042603} which penalize erroneous outputs less if they could be due to label noise. Others perform simultaneous edge detection and alignment~\cite{Yu2018SimultaneousEA} which can handle small displacements in an unsupervised manner. The task of aligning OSM annotations has already been tackled in~\cite{8518711}, using convolutional neural networks for building segmentation and a Markov Random Field for aligning buildings onto the building segmentation image. However, the neural network has to be trained on a small dataset of building image with corresponding good ground-truth annotations.

\begin{figure}
	\centering
	\includegraphics[width=1\linewidth]{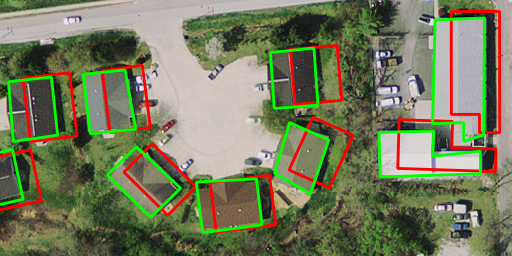}
	\caption{Qualitative alignment results on a crop of an image of Bloomington from the Inria dataset. \textcolor{red}{Red: initial OSM annotations}; \textcolor{green}{green: aligned annotations}.}
	\label{fig:qualitative_results_small}
\end{figure}

We propose in this paper to use the self-supervised multi-task multi-resolution deep learning method for aligning cadaster maps to images of~\cite{girardACCV2018} in the noisy supervision setting. The dataset used for training that method has misalignment noise and still the model learned to align. We will explore this interesting behavior and experiment a kind of unsupervised learning to correct noisy annotations with a model trained on these noisy annotations. See Fig.~\ref{fig:qualitative_results_small} for an example of results. It leverages the natural tendency of neural networks to be robust to a certain amount of noise and does not require any special loss function.

\section{Methodology}

\begin{figure*}
	\centering
	\includegraphics[width=0.7\linewidth]{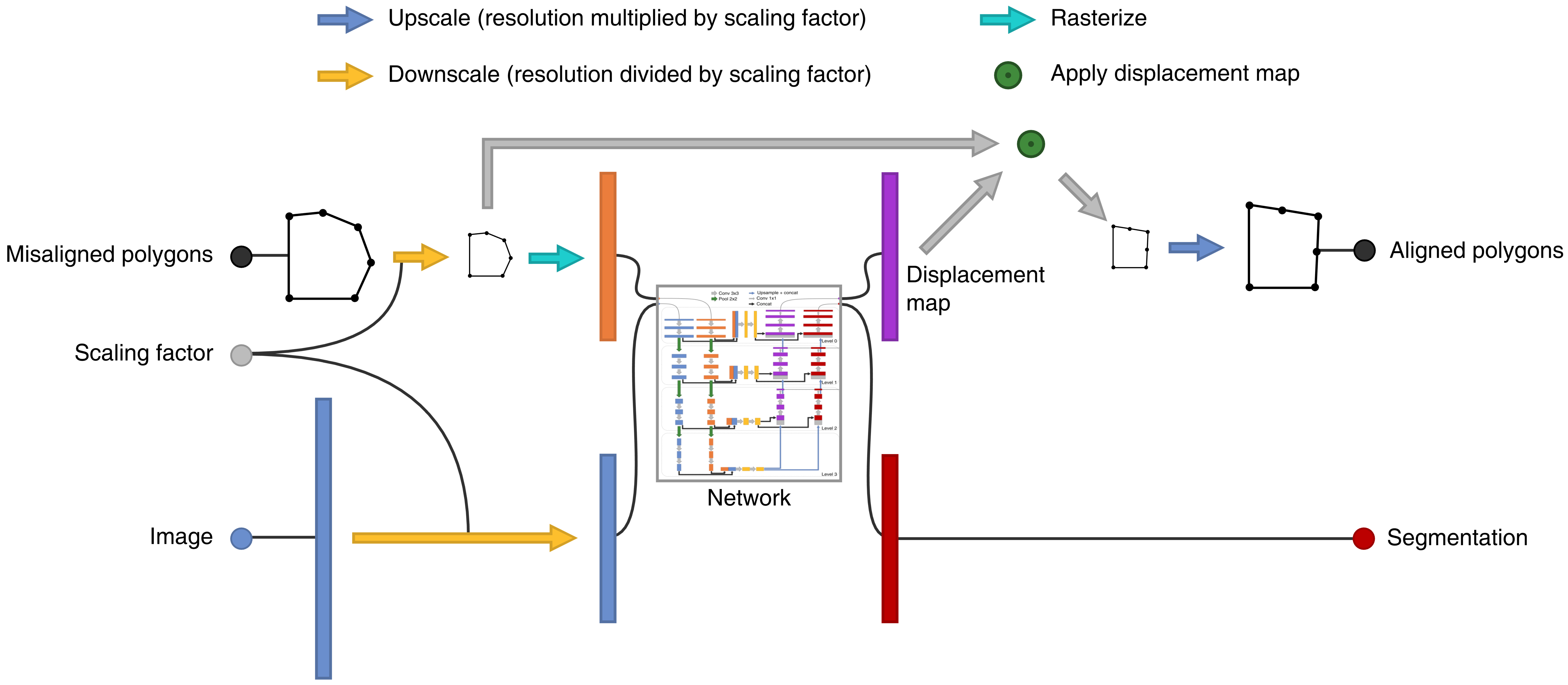}
	\vspace{-2mm}
	\caption{One step of the base alignment method, applied repeatedly at increasing resolutions for the final alignment.}
	\label{fig:pipeline}
\end{figure*}

We provide here a short description of the self-supervised multi-task multi-resolution deep learning alignment method of~\cite{girardACCV2018} (referred to by ``base alignment method'' from now on), focusing on the most relevant parts. Its code is available here: \url{https://github.com/Lydorn/mapalignment}.

\smallestsection{Mathematical modeling.} Given two images $I$ and $J$ of same size $H \times W$, but of different modalities, e.g.~with $I$ an RGB image (picture from a satellite) and $J$ a binary image (cadaster, indicating for each pixel whether it belongs to a building or not), the alignment problem aims at finding a deformation, i.e.~a 2D vector field $\f$ defined on the discrete image domain $[1, H] \times [1, W]$, such that the warped second image $J \circ (\mathrm{Id} + \f)$ is well registered with the first image $I$. To do this, in a machine learning setting, we consider triplets ($I, J, \f_\gt$) consisting of two images together with the associated ground-truth deformation $\f_\gt$. Image pairs ($I, J$) are given as inputs, and the model's estimated deformation $\fh$ is optimized to be close to the ground truth deformation $\f_\gt$.

\smallestsection{Displacement map cost function.} The displacement map loss function is the mean squared error between the predicted displacement map $\fh$ and the ground truth displacement map $\f_\gt$. The actual loss used by the base alignment method is a little more complex but for the purpose of this paper we can consider the simplified loss:
\begin{equation}
L^{\mathrm{disp}}(\fh) \; = \;\sum_{\mathbf{x} \in [1, H] \times [1, W]} \Big\| \fh(\x) - \f_\gt(\x) \Big\|_2^2
\end{equation}

\smallestsection{Model.} The neural network used by the base alignment method is a transformed U-Net~\cite{unet} with 2 image inputs and 2 image-like outputs for the displacement map and the segmentation image, see Fig.~\ref{fig:pipeline} for a schema of the model. The segmentation output is only used during training, having its own cross-entropy loss function.
The input image $I$ has 3 channels, with real values normalized to [$-1$, $1$], standing for RGB. 
The input misaligned polygon raster $J$ also has 3 channels, with Boolean values in $\{0, 1\}$, corresponding to polygon interior, edge and vertices. 
The output displacement map has 2 channels with real values in [$-4$ px, $4$ px], standing for the $x$ and $y$ components of the displacement vector.
The model uses a multi-resolution approach by applying a neural network at increasing resolutions, iteratively aligning polygons from a coarse to fine scale. The scales used are $\frac{1}{8}$, $\frac{1}{4}$, $\frac{1}{2}$ and $1$.
Thus displacements of up to $32$ px can be handled.


\section{Experimental setup}

\smallestsection{Dataset.} The model is trained on a building cadaster dataset consisting of the two available following ones: Inria Aerial Image Labeling Dataset \cite{maggiori2017dataset}, and ``Aerial imagery object identification dataset for building and road detection, and building height estimation'' \cite{bradbury_buildings_roads_height_dataset}.
The resulting dataset has 386 images (with the majority being $5000\times5000$~px) of 16 cities from Europe and USA. Each image has in average a few thousand buildings. The building footprints were pulled from OSM for all images. these polygon annotations are inconsistent across images alignment-wise (see Fig.~\ref{fig:qualitative_results_small} and ~\ref{fig:qualitative_results}). Some are perfect, and some are misaligned by up to 30~px. However the base alignment method~\cite{girardACCV2018} assumes perfect annotations in its formulation.

\smallestsection{Self-supervised training.} The model needs varied ground truth labels (displacements) in order to learn, while the dataset is assumed to be made of aligned image pairs only ($\f = 0$). The dataset is thus enhanced by adding random deformations in the form of 2D Gaussian random fields for each coordinate with a maximum absolute displacement of 32~px. The polygon annotations $A$ are then inversely displaced by the generated displacements, to compute the misaligned polygons, which are then rasterized. We obtain training triplets of the form $(I, J, \f)$ with $J = rast(A \circ (\mathrm{Id} + \f)^{-1})$. For the multi-resolution pipeline, 4 different models are trained independently with downscaling factors 8, 4, 2 and 1 (one per resolution). For clarification, labels are the ground truth displacements at each pixel, i.e.~a 2D vector, and annotations are the polygons outlining objects.

\smallestsection{Multiple-rounds training.} We train the base alignment method with the same hyper-parameters as~\cite{girardACCV2018}, which were selected to avoid overfitting. We perform multiple rounds of training on the whole dataset to achieve our goal of aligning the whole dataset. It consist of iteratively alternating between training the alignment model on the available dataset (see Alg.~\ref{alg:alignment_training}) and correcting the alignment of the training dataset to provide a better ground truth for the next training round. The multiple-rounds training is explained in Alg.~\ref{alg:multiple_round_training}.

\begin{algorithm}[h]
	\KwIn{Images $\mathcal{I} = \{I, ...\}$ and corresponding annotations $\mathcal{A} = \{A, ...\}$}
	Build dataset with random deformations: $\mathcal{D} = \{(I, J_{rand}, \f_{rand}), ...\}$ with $J_{rand} = rast(A \circ (\mathrm{Id} + \f_{rand})^{-1})$\;
	Train multi-resolution model $M$ to perform this mapping: $(I, J_{rand})\mapsto \f_{rand}$\;
	\KwOut{Trained model $M$}
	\caption{{\bf Alignment training \cite{girardACCV2018}} \label{alg:alignment_training}}
\end{algorithm}
\vspace{-5mm}
\begin{algorithm}[h]
	\KwIn{Original annotations $\mathcal{A}_0$, number of rounds $R$}
	\For{$r = 1$ \bf to $R$}{
		1. Get model $M_r$ using Alg.~\ref{alg:alignment_training} with input $\mathcal{A} = \mathcal{A}_{r-1}$\;
		2. Apply $M_r$ on the original annotations $\mathcal{A}_0$: $\mathcal{A}_r = M_r(\mathcal{A}_0)$\;
	 }
 	\KwOut{Aligned annotations $A_R$}
	\caption{{\bf Multiple-rounds training} \label{alg:multiple_round_training}}
\end{algorithm}

\smallestsection{Ablation studies.} To justify the design choices of the multiple-rounds training, we performed ablation studies. The first ablation study (AS1) changes the second step of Alg.~\ref{alg:multiple_round_training} by applying the model on the previous corrected annotations instead of the original annotations: $\mathcal{A}_r = M_r(\mathcal{A}_{r-1})$ in order to test whether it is better to iteratively align annotations. The second ablation study (AS2) trains the model only once on the original annotations, and applies it $R$ times to iteratively align the annotations (as in AS1). This is implemented by additionally replacing step~1 of Alg.~\ref{alg:multiple_round_training} by $M_r = M_1$ for $r > 1$ and leaving it as is for $r = 1$. This is meant to test the usefulness of re-training.

\smallestsection{Robustness to noise.} In an additional experiment (Noisier) we misaligned all original annotations further with random zero-mean displacements up to 16~px. We then applied our alignment method for correcting these noisier annotations to study its robustness to more noise.


\section{Results}

\begin{figure}
	\includegraphics[width=1\linewidth]{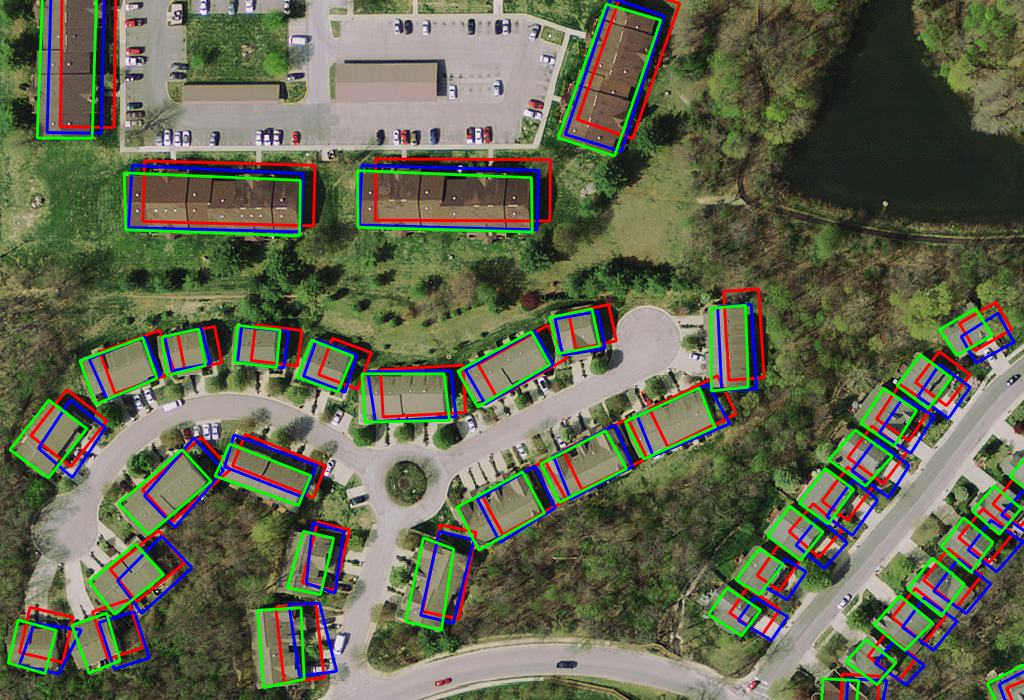}
	\caption{Qualitative alignment results on a crop of bloomington22 from the Inria dataset. \textcolor{red}{Red: initial dataset annotations}; \textcolor{blue}{blue: aligned annotations round 1}; \textcolor{green}{green:  aligned annotations round 2}.}
	\label{fig:qualitative_results}
\end{figure}

\begin{figure}
	\centering
	\vspace{-3mm}
	\includegraphics[width=1\linewidth]{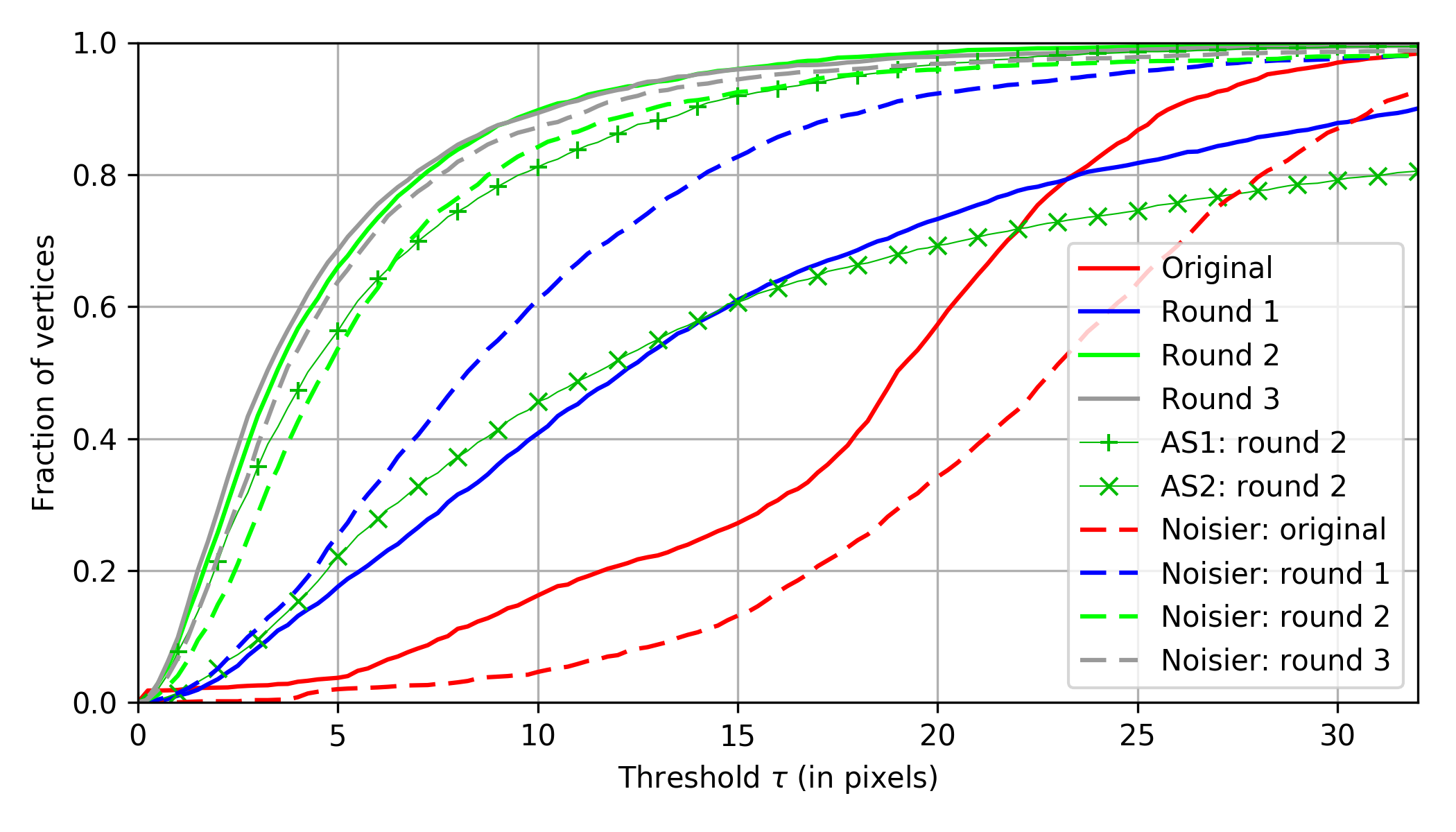}
	\vspace{-6mm}
	\caption{Accuracy cumulative distributions measured with the manually-aligned annotations of bloomington22 from the Inria dataset.}
	\label{fig:accuracies}
\end{figure}

As annotations of our dataset are noisy they cannot be used as ground truth to measure quantitative results. We can first visualize qualitative results in Fig.~\ref{fig:qualitative_results}. In order to measure the effectiveness of the multiple rounds training, we manually aligned annotations for one $5000\times5000$~px image (771 buildings) to get a good ground-truth. We chose the bloomington22 image because it has severe misalignment. To measure the accuracy of an alignment, for any threshold $\tau$ we compute the fraction of vertices whose ground truth point distance is less than $\tau$. In other words, we compute the Euclidean distance in pixels between ground truth vertices and aligned vertices, and plot the cumulative distribution of these distances in Fig.~\ref{fig:accuracies} (higher is better) for all experiments and rounds.

\section{Discussion and Conclusion}

After the first round of training, the annotations are on average better aligned than the original annotations, but in some cases the polygons are pushed into the wrong direction, resulting in poorer accuracy for some threshold levels (see the blue curve sometimes under the red curve in Fig.~\ref{fig:accuracies}). However after the second round of training, the annotation alignment has been significantly improved upon the first round (error divided by more than 3 for any quantile, cf.~green curve compared to blue curve). The 3rd round does not bring any significant improvement in this case.

\begin{figure}
	\begin{subfigure}{.67\linewidth}
		\includegraphics[height=113px]{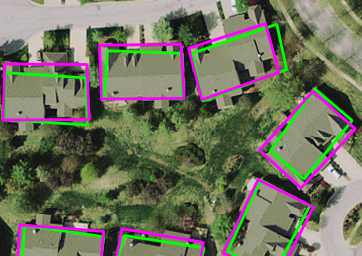}
	\end{subfigure}%
	\begin{subfigure}{.33\linewidth}
		\includegraphics[height=113px]{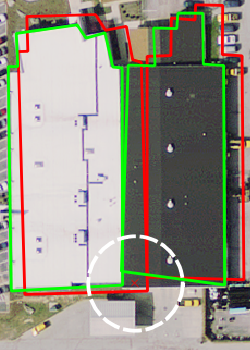}
	\end{subfigure}
	\caption{\textbf{Left}: ambiguity of the perfect ground truth annotations. \textbf{Right}: alignment failure case. \textcolor{magenta}{Magenta: manually aligned annotations}; \textcolor{red}{red: original dataset annotations}; \textcolor{green}{green: aligned annotations round 2}.}
	\label{fig:failure_cases}
\end{figure}

Note that a perfect alignment score cannot be expected, because of the ambiguity of the ``perfect'' ground truth. Indeed, when manually aligning bloomington22's annotations, we observed that the majority of buildings are annotated by a coarse polygon that does not outline the building precisely.
Best aligning such a coarse polygon to a real, more complex building becomes an ill-posed problem, with multiple equally-good solutions, which creates 
ground truth ambiguity.
See Fig.~\ref{fig:failure_cases} (left) for an illustration of this problem, especially the building on the top-right.
Fig.~\ref{fig:failure_cases} (right) shows an example of a mistake of our approach. The left building was successfully aligned (through a slight vertical and horizontal squashing), but the adjoining building on the right was not, because the model only learned smooth displacement maps. A more well-designed displacement map generation allowing discontinuities could solve such problems.

The first ablation study shows the importance of aligning the original annotations in the second step of Alg.~\ref{alg:multiple_round_training} as it achieves better accuracy. Indeed the aligned annotations after round 1 can be worse than the original annotations (see some blue polygons of Fig.~\ref{fig:qualitative_results} and the blue curve of Fig.~\ref{fig:accuracies}), and consequently more difficult to align. The second ablation study shows that the re-training step in round 2 is very important, as skipping it does not improve the alignment compared to round 1.

An explanation of how this method is able to align misaligned annotations by training on these misaligned annotations could be that the dataset contains enough perfect ground truth annotations to steer the gradient descent in the right direction, while being mildly affected by noisy labels (even if the noise is not zero-mean) if overfitting is avoided. However the last experiment (Noisier) invalidates this explanation because in that case the fraction of well-aligned ground truth is negligible and still the model was able to align noisier annotations virtually as well as it did original annotations (it however needs a 3rd round to do so). Our current tentative explanation is that ground truth labels have a zero-mean noise (without bias). For the alignment task, the network tries to minimize the average error it makes. As such it tends to predict the mean value of the labels when it cannot do better. This is the case if the label noise is independent of the input, and if overfitting noisy labels is avoided. The network will learn the mean alignment, which corresponds to the underlying perfect ground truth. This explanation is further supported by a recent work on image restoration without clean data \cite{Noise2Noise}, where noisy images are de-noised by training a de-noising network on noisy images only.

In conclusion, even noisy/misaligned annotations are useful. Our model can be iteratively trained on them and align these annotations through a multiple-round training scheme.

%
%

\begin{spacing}{0.96}
\bibliographystyle{IEEEbib}
\bibliography{biblio}

\begin{thebibliography}{1}

\bibitem{osm}
{OpenStreetMap contributors},
\newblock ``{Planet dump retrieved from https://planet.osm.org },'' 2017.

\bibitem{Mnih:2012:LLA:3042573.3042603}
Volodymyr Mnih and Geoffrey Hinton,
\newblock ``Learning to label aerial images from noisy data,''
\newblock in {\em ICML}, 2012.

\bibitem{Yu2018SimultaneousEA}
Zhiding Yu, Weiyang Liu, Yang Zou, Chen Feng, Srikumar Ramalingam, B.~V.
  K.~Vijaya Kumar, and Jan Kautz,
\newblock ``Simultaneous edge alignment and learning,''
\newblock in {\em ECCV}, Sept 2018.

\bibitem{8518711}
J.~E. Vargas-Muñoz, D.~Marcos, S.~Lobry, J.~A. dos Santos, A.~X. Falcão, and
  D.~Tuia,
\newblock ``Correcting misaligned rural building annotations in open street map
  using convolutional neural networks evidence,''
\newblock in {\em IGARSS}, 2018.

\bibitem{girardACCV2018}
Nicolas Girard, Guillaume Charpiat, and Yuliya Tarabalka,
\newblock ``Aligning and updating cadaster maps with aerial images by
  multi-task, multi-resolution deep learning,''
\newblock in {\em ACCV}, Dec 2018.

\bibitem{unet}
Olaf Ronneberger, Philipp Fischer, and Thomas Brox,
\newblock ``U-net: Convolutional networks for biomedical image segmentation,''
\newblock {\em CoRR}, 2015.

\bibitem{maggiori2017dataset}
Emmanuel Maggiori, Yuliya Tarabalka, Guillaume Charpiat, and Pierre Alliez,
\newblock ``Can semantic labeling methods generalize to any city? the
  \textsc{I}nria aerial image labeling benchmark,''
\newblock in {\em IGARSS}, 2017.

\bibitem{bradbury_buildings_roads_height_dataset}
Kyle Bradbury, Benjamin Brigman, Leslie Collins, Timothy Johnson, Sebastian
  Lin, Richard Newell, Sophia Park, Sunith Suresh, Hoel Wiesner, and Yue Xi,
\newblock ``Aerial imagery object identification dataset for building and road
  detection, and building height estimation,'' July 2016.

\bibitem{Noise2Noise}
Jaakko Lehtinen, Jacob Munkberg, Jon Hasselgren, Samuli Laine, Tero Karras,
  Miika Aittala, and Timo Aila,
\newblock ``Noise2noise: Learning image restoration without clean data,''
\newblock {\em CoRR}, 2018.

\end{thebibliography}
\end{spacing}

\end{document}